\documentclass{article}
\usepackage{xcolor}
\usepackage{booktabs}
\usepackage{spconf,amsmath,graphicx,dirtytalk,enumitem,epstopdf,setspace}
\usepackage{authblk}
\usepackage{tabularx}

\title{A Chat About Boring Problems: Studying GPT-based text normalization}
\name{Yang Zhang$^{*}$, Travis M. Bartley$^{\dagger}$\sthanks{Equal contribution}, Mariana Graterol-Fuenmayor, Vitaly Lavrukhin, Evelina Bakhturina, Boris Ginsburg}

\address{Nvidia Corporation, 
\\ City University of New York, Graduate Center$^{\dagger}$}

\begin{document}
\maketitle
\begin{abstract}
Text normalization - the conversion of text from written to spoken form - is traditionally assumed to be an ill-formed task for language modeling. In this work, we argue otherwise. We empirically show the capacity of Large-Language Models (LLM) for text normalization in few-shot scenarios. Combining self-consistency reasoning with linguistic-informed prompt engineering, we find LLM-based text normalization to achieve error rates approximately 40\% lower than production-level normalization systems. Further, upon error analysis, we note key limitations in the conventional design of text normalization tasks. We create a new taxonomy of text normalization errors and apply it to results from GPT-3.5-Turbo and GPT-4.0. Through this new framework, we identify strengths and weaknesses of LLM-based TN, opening opportunities for future work.   
\end{abstract}

\section{Introduction}

\label{sec:intro}
\begin{figure}[h]
  \centering
  \includegraphics[width=\linewidth]{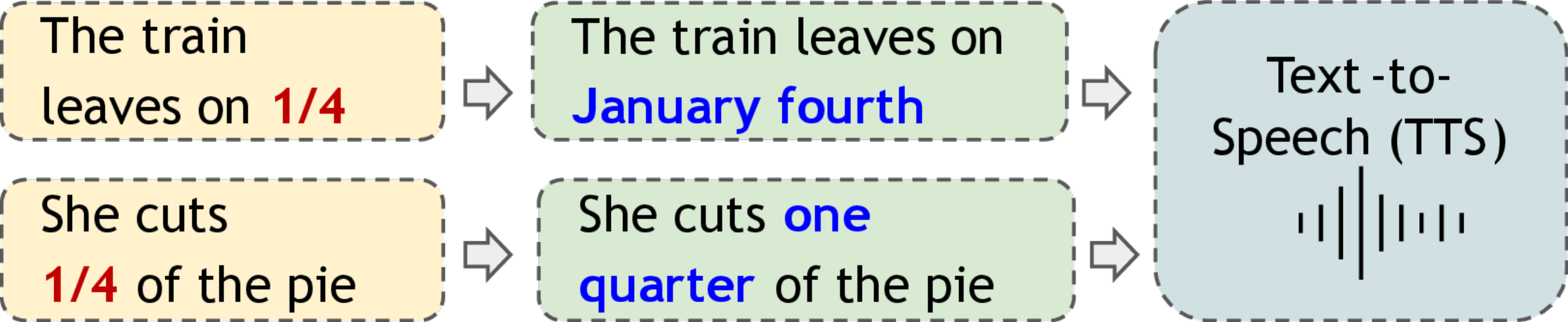}
  \caption{Example of text normalization across semiotic classes. The normalization of the string ("1/4") varies given semantic context.}
  \label{fig:context_normalization_examples}
\end{figure}

Text-normalization (TN) - the conversion of text symbols to orthographic spelling - is an essential prepossessing step for text-to-speech generation (TTS). Used to process numeric and abbreviated strings, TN is distinct from other prepossessing tasks (e.g. grapheme-to-phoneme transcription) due to its dependence on sentential context. For example, consider the normalization of "1/4" as seen in Figure~\ref{fig:context_normalization_examples}. Without context, the intended pronunciation is ambiguous. A naive TTS model could produce "January the fourth", "April first", "one fourth", or simply "one slash four". Yet only the first would be valid as the date of a U.S. locale. Such cases are prevalent across TN inputs and require special care to improve downstream systems.

Historically, researchers addressed TN with weighted finite-state grammars (WFST)~\cite{ebden2015kestrel,automata,sproat2001normalization,roark-etal-2012-opengrm}. Graph representations of complex text relations, TN grammars are notoriously labor intensive and require specialized knowledge to develop. As such, there has long been interest in adopting probabilistic methods in their place \cite{sproat2001normalization}. Recent advances in neural architectures have led to renewed focus, with some works attempting to use mixed-methods to replicate WFST behavior \cite{shallowfusion, tyagi-etal-2021-proteno}. Others have approached TN as a seq2seq task akin to machine translation \cite{Sproat2017,mansfield-etal-2019-neural,zhang2019neural, jiang2021}. Though these TN systems achieve good performance, they have seen limited adoption. 

This stems from concern for \textit{unrecoverable errors}: erroneous outputs that prevent inference of original text (e.g. output of "three hundred and fifty six" for an input of "556"). Since TN domains often include low fault-tolerant environments (e.g. Financial quantities, Medical results), it is argued that only deterministic, rule-based methods are sufficiently reliable for end-user development. As such, most production level systems rely on hand-coded weighted finite-state grammars in all \cite{ebden2015kestrel,zhang21ja_interspeech} or some \cite{pusateri17_interspeech} capacity. These limitations have led Richard Sproat – one of the pioneers of modern TN - to claim, "...until one can solve 'boring' problems like this using purely AI methods, one cannot claim that AI is a success...so far nobody has eliminated the problem of unrecoverable errors" ~\cite{sproat-2022-boring}.

In this work, we respond to Sproat: \textit{true} unrecoverable errors are minimal for neural TN. Indeed, reliable TN is achievable with common application of large-language models (LLM). The perceived extent of unrecoverable errors instead lies in TN error documentation. To demonstrate, we introduce a granular label schema for TN errors, along with a simple labeling tool for analysis via the NeMo Text-Processing toolkit~\cite{zhang2021nemo}. Using  GPT-3.5-Turbo and GPT-4.0, we evaluate TN performance over few-shot prompting and consistency reasoning scenarios. We find LLM based TN (LLM-TN) to outperforms production-level WFST systems by around 40\% percent. Further, LLM-TN errors prove near non-existent, with previous concerns overshadowing LLM-TN's capability to address limitations of WFSTs.  

\section{Method}
\subsection{Error Taxonomy}
\label{sec:taxonomy}
We first motivate our rationale for a TN error taxonomy: As defined, TN does not assume an injective relation. In Figure~\ref{fig:context_normalization_examples}, the date "1/4" may be verbalized as "January the fourth" or "the fourth of January" without changing meaning. Yet, the semantically equivalent "the day after January the third" would be perceived as incorrect. 

This relates to the linguistic notion of \textit{felicity}. Broadly, felicitous statements are the subset of grammatical sentences that are contextually permissible (with \textit{infelicitous} statements being the disjoint) \cite{austin}. While in specific contexts only one sentence may be felicitous, other contexts permit many equivalent substitutions. 

Consider TN for a US telephone number: \\

\textbf{Text}: 312-236-2012 

\textbf{a}: three twelve two thirty six twenty twelve 

\textbf{b}: three one two two three six two zero one two \\

Both strings are felicitous  in regards to the intended context of U.S. phone-numbers. Yet, were \textbf{a} our gold example, \textbf{b} would be treated as an error. As such, the failure to account for felicity at the sentence level produces a false negative. An effective analysis of TN should account for this variation. In contrast, consider: \\

\textbf{Text}: b. 03-02-2001

\textbf{c}: born march second two thousand one

\textbf{d}: b zero three zero two two thousand one \\

Absent context, both \textbf{c} and \textbf{d} are valid normalizations, However, only \textbf{c} is acceptable for a date of birth and thus \textbf{d} is infelicitous. Yet, \textbf{d} still retains the original textual information and is preferable to other false productions (i.e. it is not unrecoverable). A proper TN analysis would note this distinction.

The above reasoning motivates error categories for TN analysis. They distinguish infelicitous errors from unrecoverable ones. Further, they delineate causes stemming from extra-TN behavior (e.g. noisy text-correction, translation of foreign words, hallucinations) that can induce both unrecoverable errors and felicitous alternative outputs. In total, our analysis identifies six TN error categories. They are:

%\break
\begin{itemize}[noitemsep]
    \item \textbf{format}: Infelicitous outputs (as described above). 
    \item \textbf{paraphrase}: Text replacement, deletion, or reordering. (e.g. "happily he danced" $\rightarrow$ "he danced happily.")
    \item \textbf{fix}: Grammar or spelling correction of a noisy sample. (e.g. "he sleep yesterday" $\rightarrow$ "he slept yesterday".)
    \item \textbf{artifact}: Introduction of non-original text due to idiosyncratic LLM behavior. (e.g. Prepending "Sure, I can do that".)
    \item \textbf{translation}: Normalization of non-English text. (e.g. "Louis XIV était roi de france" $\rightarrow$ "Louis quatorze était roi de france".) 
    \item \textbf{other}: Miscellaneous category.
\end{itemize}

As unrecoverability can stem from multiple factors, only the \textit{format} label excludes unrecoverable errors. All other label categories require manual evaluation to determine if predictions are unrecoverable or simply alternate felicitous outputs.

\begin{figure*}[h]
  \centering
  \includegraphics[width=0.99\textwidth,trim=150 140 50 140,clip]{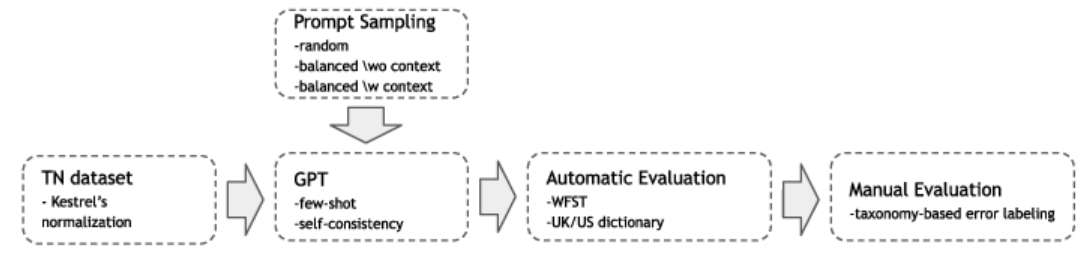}
  \caption{GPT normalization and evaluation pipeline.}
  \label{fig:pipeline}
\end{figure*}

\subsection{Experiments}
To demonstrate our taxonomy's utility, we compare performance of neural TN architecture against WFST methods. We note the strongest evidence for neural-based TN would come from the efficacy of a general purpose model. That is, if our taxonomy shows high performance on a \textit{non-specialized} model, then we may extrapolate improved performance for TN-exclusive models. As such, we compare against WFST baselines with an LLM and standard prompting techniques, leaving comparison of other neural methods for future work.

For our models, we compare GPT-3.5-Turbo and GPT-4.0 \cite{brown2020language,openai2023gpt4} (March 1st and current release, respectively) against Google's Kestrel WFST \cite{ebden2015kestrel}. We seed the LLM with sample conversation turns of the format $Normalize: \{INPUT\}$ followed by a target normalization. Though we initially experimented with variations of the TN prompt, we found no substantial difference in output quality. As such, we retained the minimal prompt to reduce rate-limiting by the GPT API.

To examine influence of coverage on TN quality, we randomly sample sentence-normalization pairs until each samples covers at least one of the common TN domains (dates, time, addresses, URLs, digit strings, fractional amounts, telephone numbers, currency, measures, decimal strings, cardinal numbers, ordinal numbers, letters, verbatim text, and plain text). To explore influence of sentential context on TN quality, we create pseudo-sentences that randomly concatenate phrasal substrings until all TN domains are covered. For baseline LLM-TN performance, we also prompt with randomly selected sentences.

For each sampling method, we prompt with sufficient examples such that $r=1,2,3$ occurrences of each TN domain are present across totality of samples. For original sentence (w/ context) sampling, this produces 14, 28, and 40 example sentences of 877, 1646, and 2381 respective tokens. For pseudo-sentence sampling, this produces 12, 24, and 36 sentences of 1010, 1897, 2789 respective. For random sampling, we sample the equivalent number of sentences as w/ context r=1, 2, 3, with 542, 1142, and 1586 respective tokens.\footnote{Token sizes are calculated using OpenAI's token estimation tool https://platform.openai.com/tokenizer.}

For sampling experiments, we evaluate only with GPT-3.5-Turbo (due to rate limitations for GPT-4.0 API).

\subsection{Dataset}
For our experiments, we use the Google TN dataset~\cite{Sproat2017}. Comprised of around 1.1 billion words from Wikipedia, it is one of the few publicly available English TN datasets and provides sizeable coverage of common TN domains. Individual tokens are paired with domain tags (see \cite{Sproat2017, taylor_tts, esch17_interspeech} for common TN domains) and a target normalization generated with Google's Kestrel TN system~\cite{ebden2015kestrel}.

As Kestrel is designed for TTS, the dataset includes several artifacts such as "sil" and special suffixes for single character. We remove these as a prepossessing step. As well, we filter out non-English text when present.

For sampling experiments, we gather examples from the provided training set and evaluate over 1000 sentences from the provided evaluation set. Final performance is evaluated over the entire test set, comprised of 7551 sentences. 

\section{Results}
\label{sec:results}
\subsection{Sampling Experiments}
For evaluation, we filter out irrelevant variations between target and prediction samples. We remove common GPT insertions ("No need for normalization", "This sentence seems to be incomplete") and filter obvious non-English samples using regex search. Simple felicitous variations (such as "August fourth" vs. "fourth of August") are corrected using a WFST-grammar. We also correct UK/US spelling differences with a pre-defined dictionary. Finally, we screen for variations in punctuation and article insertion. 

Table~\ref{tab:prompt} shows GPT-3.5-Turbo's exact accuracy on the development subset before labeling. Sampling two in context examples per semiotic class yields the highest accuracy of 91.1\% when compared against target predictions.

To improve performance, we augment the three best performing configurations with self-consistency reasoning ~\cite{wang2022self}. As we seek to demonstrate efficacy of standard LLM methods, we only use the same recommended sampling parameters of $t=0.5$ and majority voting \cite{wang2022self,radford2019language,holtzman2019curious}. However, for tractability we poll 20 completion outputs instead of 40. (see Figure~\ref{fig:pipeline}). Results from self-consistency is provided in Table~\ref{tab:prompt}. For all configurations, self-consistency leads to noticeable performance gains.

\begin{table}
\caption{\% TN accuracy of GPT-3.5-Turbo over first 1000 samples in dev set. $r$ indicates number of examples per TN domain. Second number (if provided) is \% accuracy after self-consistency augmentation (N=20; t=0.5).}
\begin{tabular}{l|lll}
Prompt Samp. Method & r=1 & r=2 & r=3\\ \hline
w/ context  & 88.8/91.1     & \textbf{91.1/92.1} & 90.5     \\
w/o context & 89.8     & 90.4/91.5          & 89.8 \\
Rand.   & 87.1     & 88.7          & 88.9     \\
\end{tabular}
\label{tab:prompt}
\end{table}

% \begin{figure}[h]
%   \centering
%   \includegraphics[width=\linewidth]{pics/error_labeling.png}
%   \caption{HTML error labeling tool. "Source" is written form, "target" is Kestrel's normalization and "predict" is GPT's normalization.}
%   \label{fig:tool}
% \end{figure}

\subsection{Final Performance}

\begin{table}
\caption{TN error counts by label type for GPT-3.5-Turbo, GPT-4.0, and Kestrel. Errors '(Post Auto)' and 'Errors (Post Manual)' produced from automatic and manual evaluation, respectively. Kestrel errors are target normalizations differing from GPT-4.0 predictions.}
\label{tab:errortypes}
\begin{tabular}{l|r|r||r}
Error Type       & GPT-3.5 & GPT-4.0 & Kestrel  \\ \hline
Errors (Post Auto)    & 692            & 540      &   540*      \\
Errors (Post Manual)   & 264            & 108      &   174    \\
Foreign Language & 29             & 12       &   26     \\
Format           & 173            & 64       &   143    \\
Paraphrase       & 26             & 12       &   1      \\
Fix              & 2              & 6        &   0      \\
Artifact         & 19             & 10        &   3      \\
Other            & 15             & 4        &   0      \\ \hline
Final Accuracy   & .965           & \textbf{.986}     &   .977 \\
Unrecoverable Errors &  N/A         & 9        &   1 
\end{tabular}
\end{table}

Using the best performing self-consistency configuration (w/ context; r=2), we evaluate both GPT models over the entire test set. Further, we perform manual evaluation with our error taxonomy as described in Section~\ref{sec:taxonomy}. Felicitous predictions are relabeled, while infelicitous and unrecoverable errors are recorded in Table~\ref{tab:errortypes}. We restrict unrecoverable errors to only English strings and ignore non-English script and diacritics (e.g. Greek characters, accent marks).

From GPT-3.5-Turbo's 692 errors after automatic evaluation, manual labeling found only 264 to be actual errors (see Section~\ref{sec:taxonomy}). GPT-4.0 significantly improves results, with its initial 540 errors reducing to 108 errors. The 540 test cases where GPT-4.0 differs from Kestrel after automatic evaluation are also manually analyzed before comparison with GPT-4.0 errors. After evaluation, we found GPT-4.0 makes approximately 40\% less errors than Kestrel. Of these errors, over half were simply infelicitous normalizations. Regarding unrecoverable errors, GPT-4.0 made only 9 across the entire dataset, representing around 0.1\% of outputs.

\section{Discussion}
We observe self-consistency and domain coverage are critical for performance. Interestingly, improvements from sample-size appears to saturate once prompts exceed 2000 tokens. We surmise this stems from 'lost in the middle' phenomenon~\cite{liu2023lost}: context knowledge for normalization may be lost over critically large token spans. 

An error taxonomy is necessary for thorough TN analysis. Among all models, infelicitous normalizations (\textit{format}) were the majority of errors. Unrecoverable errors were near non-existent for GPT-4.0. While their frequency is one order of magnitude above Kestrel's errors, this overshadows improvements LLM-TN makes over WFST. During manual labeling, we found several supposed errors came from contamination in test data that GPT-4.0 had corrected. In fact, Kestrel's one unrecoverable error was corrected by the GPT model (see Table \ref{tab:gpt_errors}).

Further, egregious unrecoverable errors (e.g. hallucination of an extra digit) are a small amount of errors (we count three on first pass). Other unrecoverable errors (like the final two in Table \ref{tab:gpt_errors}) still allow some level of inference of original text. For instance, the penultimate example ("in 1893-94") still preserves the root digit values. Meanwhile the final prediction's insertion of "crore" indicates a quantity prominent in South Asian counting systems. 

Given the improvements from self-consistency, low rate of error, and the minor nature of many unrecoverable errors, we believe that the remaining limitation for LLM-TN is disambiguation of TN outputs. That is, while prompts are sufficient to induce TN behavior in the LLM, the variety of potential TN outputs are difficult to disambiguate. Improvements can likely be made by addition of some explicit filtering mechanisms.

Such alterations would be similar to hybrid TN architectures \cite{shallowfusion, tyagi-etal-2021-proteno,pusateri17_interspeech}. Such architectures use domain categories for TN substrings and limit specific outputs per category. Given the small error rate of LLM-TN, we believe the minor addition of some of these constraints could make unrecoverable errors non-existent.

\begin{table}
\caption{Examples of GPT-4.0 TN errors with error labels.}
\begin{tabular}{@{} l @{}}
\toprule
\textbf{Input:} 2007 i triple e conference \\
\textbf{GPT [paraphrase]:} two o o seven i {\color{red}e e e} conference \\
\textbf{Target:} two o o seven i triple e conference. \\
\midrule
\textbf{Input:} 27 oct . 2010 : 8 \\
\textbf{GPT [format]:} twenty seventh { \color{red} oct .} two thousand ten : eight \\
\textbf{Target:} the twenty seventh of october two thousand ten : eight \\
\midrule
\textbf{Input/target:} it was originally mooted \\
\textbf{GPT[paraphrase]:} it was originally {\color{red}suggested} \\
\midrule
\textbf{Input:} cd 004913 . \\
\textbf{GPT [unrecov.]:} c d {\color{red} four zero } four nine one three . \\
\textbf{Target:} c d o o four nine one three . \\
\midrule
\textbf{Input:} student stories of 9/11 \\
\textbf{GPT [correct]:} student stories of {\color{teal} nine eleven}  \\
\textbf{Target [unrecov.]:} student stories of {\color{red} nine elevenths} \\
\midrule
\textbf{Input:} in 1893 - 94 \\
\textbf{GPT [unrecov.]:} in eighteen ninety three ninety {\color{red}fourths} \\
\textbf{Target [correct]:} in eighteen ninety three - ninety four \\
\midrule
\textbf{Input [wrong]:} more than rs.10 , 00 {\color{red}, 000 , 00} \\
\textbf{GPT [unrecov.]:} more than {\color{red} ten crore} \\
\textbf{Target [wrong]:} more than ten {\color{red} rupees , o o , o o o , o o} \\
\bottomrule
\end{tabular}
\label{tab:gpt_errors}
\end{table}

\section{Conclusion}
We empirically demonstrate effectiveness of LLMs for TN tasks. After careful analysis of TN errors, we find modern LLMs can outperform rule-based normalization systems by approximately 40\%. More importantly, they perform with minuscule instances of unrecoverable normalization errors. We show that LLM-TN outputs are of high quality and can even self-correct noisy errors that cause degradation in WFST inputs.

Further, we demonstrate a need for greater nuance in TN analysis. TN outputs can cover a wide range of felicitous targets and are ill-served by binary label schemes. We encourage future work to explore refinements to this taxonomy to develop even more effective TN systems.  

\vfill\pagebreak

\bibliographystyle{IEEEbib}
\bibliography{strings,refs}

\end{document}